\def\year{2017}\relax
\documentclass[letterpaper]{article}
\usepackage{aaai17}
\usepackage{times}
\usepackage{helvet}
\usepackage{courier}
\usepackage{epsfig}
\usepackage{graphicx}
\usepackage{amsmath}
\usepackage{amssymb}
\usepackage{bm}
\usepackage{array}
\usepackage{color}

\begin{document}
%
\title{A Riemannian Network for SPD Matrix Learning}
\author{Zhiwu Huang and Luc Van Gool\\
	Computer Vision Lab, ETH Zurich, Switzerland \\
	{\tt\small \{zhiwu.huang, vangool\}@vision.ee.ethz.ch}}
\maketitle
\begin{abstract}
Symmetric Positive Definite (SPD) matrix learning methods have become popular in many image and video processing tasks, thanks to their ability to learn appropriate statistical representations while respecting Riemannian geometry of underlying SPD manifolds. In this paper we build a Riemannian network architecture to open up a new direction of SPD matrix non-linear learning in a deep model. In particular, we devise bilinear mapping layers to transform input SPD matrices to more desirable SPD matrices, exploit eigenvalue rectification layers to apply a non-linear activation function to the new SPD matrices, and design an eigenvalue logarithm layer to perform Riemannian computing on the resulting SPD matrices for regular output layers. For training the proposed deep network, we exploit a new backpropagation with a variant of stochastic gradient descent on Stiefel manifolds to update the structured connection weights and the involved SPD matrix data. We show through experiments that the proposed SPD matrix network can be simply trained and outperform existing SPD matrix learning and state-of-the-art methods in three typical visual classification tasks.
\end{abstract}

\section{Introduction}

Symmetric Positive Definite (SPD) matrices are often encountered and have made great success in a variety of areas. In medical imaging, they are commonly used in diffusion tensor magnetic resonance imaging \cite{pennect2006aid,arsigny2007led,jayasumana2013kernel}. In visual recognition, SPD matrix data provide powerful statistical representations for images and videos. Examples include region covariance matrices for pedestrian detection \cite{tuzel2006region,tuzel2008pedestrian,tosato2010region}, joint covariance descriptors for action recognition \cite{harandi2014manifold}, image set covariance matrices for face recognition \cite{wang2012covariance,huang2014cvpr,huang2015leml} and second-order pooling for object classification \cite{ionescu2015matrix}.

As a consequence, there has been a growing need to carry out effective computations to interpolate, restore, and classify SPD matrices. However, the computations on SPD matrices often accompany with the challenge of their non-Euclidean data structure that underlies a Riemannian manifold \cite{pennect2006aid,arsigny2007led}. Applying Euclidean geometry to SPD matrices directly often results in undesirable effects, such as the swelling of diffusion tensors \cite{pennect2006aid}. To address this problem, \cite{pennect2006aid,arsigny2007led,sra2011sdiv} introduced Riemannian metrics, e.g., Log-Euclidean metric \cite{arsigny2007led}, to encode Riemannian geometry of SPD manifolds properly.

By employing these well-studied Riemannian metrics, existing SPD matrix learning approaches typically flatten SPD manifolds via tangent space approximation \cite{tuzel2008pedestrian,tosato2010region,carreira2012sem,fathydiscriminative2016ijcai}, or map them into reproducing kernel Hilbert spaces \cite{harandi2012sparse,wang2012covariance,sanin2013spatio,minh2014nips,faraki2015approximate,zhang2015online}. To more faithfully respect the original Riemannian geometry, recent methods \cite{harandi2014manifold,huang2015leml} adopt a geometry-aware SPD matrix learning scheme to pursue a mapping from the original SPD manifold to another one with the same SPD structure. However, all the existing methods merely apply shallow learning, with which traditional methods are typically surpassed by recent popular deep learning methods in many contexts in artificial intelligence and visual recognition.

In the light of the successful paradigm of deep neural networks (e.g., \cite{lecun1998gradient,krizhevsky2012imagenet}) to perform non-linear computations with effective backpropagation training algorithms, we devise a deep neural network architecture, that receives SPD matrices as inputs and preserves the SPD structure across layers, for SPD matrix non-linear learning. In other words, we aim to design a deep learning architecture to non-linearly learn desirable SPD matrices on Riemannian manifolds. In summary, this paper mainly brings three innovations:
\begin{itemize}
	\item A novel Riemannian network architecture is introduced to open a new direction of SPD matrix deep non-linear learning on Riemannian manifolds.
	\item This work offers a paradigm of incorporating the Riemannian structures into deep network architectures for compressing both of the data space and the weight space.
	\item A new backpropagation is derived to train the proposed network with exploiting a stochastic gradient descent optimization algorithm on Stiefel \mbox{manifolds}.
\end{itemize}

\section{Related Work}
\label{gen_inst}

Deep neural networks have exhibited their great powers when the processed data own a Euclidean data structure. In many contexts, however, one may be faced with data defined in non-Euclidean domains. To tackle graph-structured  data, \cite{bruna2014spectral} presented a spectral formulation of convolutional networks by exploiting a notion of non shift-invariant convolution that depends on the analogy between the classical Fourier transform and the Laplace-Beltrami eigenbasis. Following \cite{bruna2014spectral}, a localized spectral network was proposed in \cite{boscaini2015learning} to non-Euclidean domains by generalizing the windowed Fourier transform to manifolds to extract the local behavior of some dense intrinsic descriptor. Similarly, \cite{masci2015geodesic} proposed a `geodesic convolution' on non-Euclidean local geodesic system of coordinates to extract `local patches' on shape manifolds. The convolutions in this approach were performed by sliding a window over the shape manifolds.

Stochastic gradient descent (SGD) has been the workhorse for optimizing deep neural networks. As an application of the chain rule, backpropagation is commonly employed to compute Euclidean gradients of objective functions, which is the key operation of SGD. Recently, the two works \cite{ionescu2015matrix,gao2016matrix} extended backpropagation directly on matrices. For example, \cite{ionescu2015matrix} formulated matrix backpropagation as a generalized chain rule mechanism for computing derivatives of composed matrix functions with respect to matrix inputs. Besides, the other family of network optimization algorithms exploits Riemannian gradients to handle weight space symmetries in neural networks. For instance, recent works \cite{bottou2010large,bonnabel2013stochastic,olivier2015a,marceau2016practical} developed several optimization algorithms by building Riemannian metrics on the activity and parameter space of neural networks, treated as Riemannian manifolds.

\begin{figure*}[t]
	\begin{center}
		\includegraphics[width=0.75\linewidth]{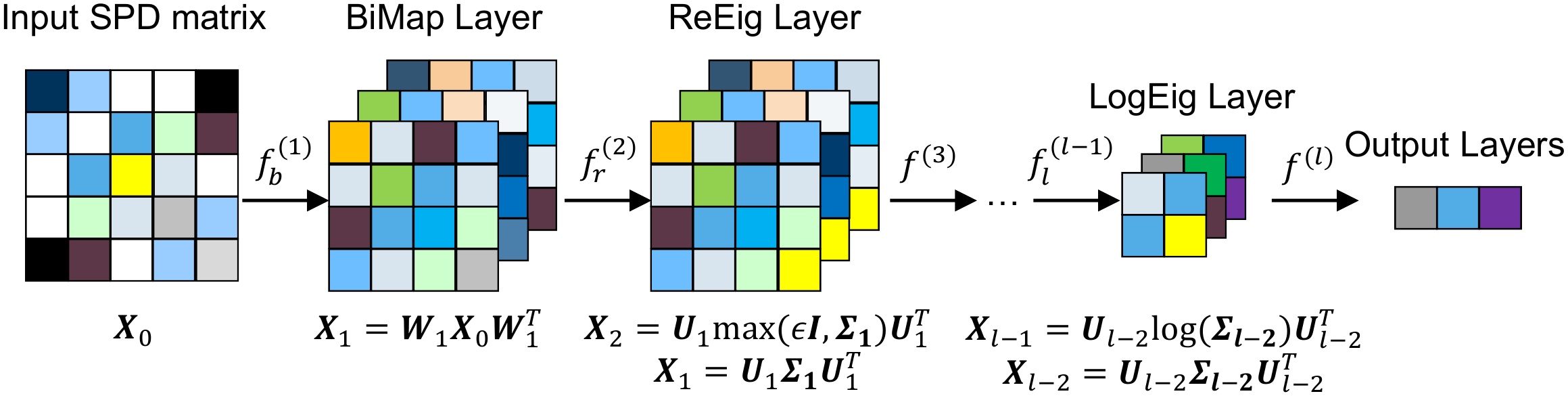}
	\end{center}
	\caption{Conceptual illustration of the proposed SPD matrix network (SPDNet) architecture.}
	\label{fig:long}
	\label{Fig1}
\end{figure*}

\section{Riemannian SPD Matrix Network}
\label{spdnet}

Analogously to the well-known convolutional network (ConvNet), the proposed SPD matrix network (SPDNet) also designs fully connected convolution-like layers and rectified linear units (ReLU)-like layers, named bilinear mapping (BiMap) layers and eigenvalue rectification (ReEig) layers respectively. In particular, following the classical manifold learning theory that learning or even preserving the original data structure can benefit classification, the BiMap layers are designed to transform the input SPD matrices, that are usually covariance matrices derived from the data, to new SPD matrices with a bilinear mapping. As the classical ReLU layers, the proposed ReEig layers introduce a non-linearity to the SPDNet by rectifying the resulting SPD matrices with a non-linear function. Since SPD matrices reside on non-Euclidean manifolds,  we have to devise an eigenvalue logarithm (LogEig) layer to carry out Riemannian computing on them to output their Euclidean forms for any regular output layers. The proposed Riemannian network is conceptually illustrated in Fig.\ref{Fig1}.

\subsection{BiMap Layer}

The primary function of the SPDNet is to generate more compact and discriminative SPD matrices.
To this end, we design the BiMap layer to transform the input SPD matrices to new SPD matrices by a bilinear mapping $f_b$ as
\begin{equation}
\bm{X}_k = f_b^{(k)}(\bm{X}_{k-1}; \bm{W}_k)=\bm{W}_k\bm{X}_{k-1}\bm{W}_k^T,
\label{Eq1}
\end{equation}
where $\bm{X}_{k-1} \in Sym^{+}_{d_{k-1}}$ is the input SPD matrix of the $k$-th layer, $\bm{W}_k  \in  \mathbb{R}_{*}^{d_{k} \times d_{k-1}}, (d_{k} < d_{k-1})$ is the transformation matrix (connection weights), $\bm{X}_k \in \mathbb{R}^{d_{k} \times d_{k}}$ is the resulting matrix. Note that multiple bilinear mappings can be also performed on each input. To ensure the output  $\bm{X}_k$ becomes a valid SPD matrix, the transformation matrix $\bm{W}_k$ is basically required to be a row full-rank matrix. By applying the BiMap layer, the inputs on the original SPD manifold $Sym^{+}_{d_{k-1}}$ are transformed to new ones which form another SPD manifold $Sym^{+}_{d_{k}}$. In other words, the data space on each BiMap layer corresponds to one SPD manifold.

Since the weight space $\mathbb{R}_{*}^{d_{k} \times d_{k-1}}$ of full-rank matrices is a non-compact Stiefel manifold where the distance function has no upper bound, directly optimizing on the manifold is infeasible. To handle this problem, one typical solution is to additionally assume the transformation matrix $\bm{W}_k$ to be orthogonal (semi-orthogonal more exactly here)  so that they reside on a compact Stiefel manifold $St(d_{k},d_{k-1})$ \footnote{A compact Stiefel manifold $St(d_{k},d_{k-1})$ is the set of $d_{k}$-dimensional orthonormal matrices of the $\mathbb{R}^{d_{k-1}}$.}. As a result, optimizing over the compact Stiefel manifolds can achieve optimal solutions of the transformation matrices.

\subsection{ReEig Layer}

In the context of ConvNets, \cite{jarrett2009best,nair2010rectified} presented various rectified linear units (ReLU) (including the $\max(0, x)$ non-linearity) to improve discriminative performance. Hence, exploiting ReLU-like layers to introduce a non-linearity to the context of the SPDNet is also necessary. Inspired by the idea of the $\max(0, x)$ non-linearity, we devise a non-linear function $f_r$ for the ReEig ($k$-th) layer to rectify the SPD matrices by tuning up their small positive eigenvalues:
\begin{equation}
\bm{X}_k = f_r^{(k)}(\bm{X}_{k-1})=\bm{U}_{k-1}\max(\epsilon\bm{I}, \bm{\Sigma}_{k-1})\bm{U}_{k-1}^T,
\label{Eq2}
\end{equation}
where $\bm{U}_{k-1}$ and $\bm{\Sigma}_{k-1}$ are achieved by eigenvalue decomposition (EIG) $\bm{X}_{k-1}=\bm{U}_{k-1}\bm{\Sigma}_{k-1}\bm{U}_{k-1}^T$, $\epsilon$ is a rectification threshold, $\bm{I}$ is an identity matrix, $\max(\epsilon\bm{I}, \bm{\Sigma}_{k-1})$ is a diagonal matrix $\bm{A}$ with diagonal elements being defined as
\begin{align}
	\bm{A}(i,i) =
	\begin{cases}
	\bm{\Sigma}_{k-1}(i,i), & \bm{\Sigma}_{k-1}(i,i) > \epsilon,\\
		\epsilon, & \bm{\Sigma}_{k-1}(i,i) \leq \epsilon.
	\end{cases}
	\label{Eq3}
\end{align}

Intuitively, Eqn.\ref{Eq2} prevents the input SPD matrices from being close to non-positive ones (while the ReLU yields sparsity). Nevertheless, it is not originally designed for regularization because the inputs are already non-singular after applying BiMap layers. In other words, we always set $\epsilon$ above the top-n smallest eigenvalue even when the eigenvalues of original SPD matrices are all much greater than zero.

Besides, there also exist other feasible strategies to derive a non-linearity on the input SPD matrices. For example, the sigmoidal function \cite{cybenko1989approximation} could be considered to extend Eqn.\ref{Eq2} to a different activation function. Due to the space limitation, we do not discuss this any further.

\subsection{LogEig Layer}

The LogEig layer is designed to perform Riemannian computing on the resulting SPD matrices for output layers with objective functions. As studied in \cite{arsigny2007led}, the Log-Euclidean Riemannian metric is able to endow the Riemannian manifold of SPD matrices with a Lie group structure so that the manifold is reduced to a flat space with the matrix logarithm operation $\log(\cdot)$ on the SPD matrices. In the flat \mbox{space}, classical Euclidean computations can be applied to the domain of SPD matrix logarithms. Formally, we employ the Riemannian computation \cite{arsigny2007led} in the $k$-th layer to define the involved function $f_l$ as
\begin{equation}
\bm{X}_{k} = f_l^{(k)}(\bm{X}_{k-1})=\log(\bm{X}_{k-1})=\bm{U}_{k-1}\log(\bm{\Sigma}_{k-1})\bm{U}_{k-1}^T,
\label{Eq4}
\end{equation}
where $\bm{X}_{k-1}=\bm{U}_{k-1}\bm{\Sigma}_{k-1}\bm{U}^T_{k-1}$ is an EIG operation, $\log(\bm{\Sigma}_{k-1})$ is the diagonal matrix of eigenvalue logarithms.

For SPD manifolds, the Log-Euclidean Riemannian computation is particularly simple to use and avoids the high expense of other Riemannian computations \cite{pennect2006aid,sra2011sdiv}, while preserving favorable theoretical properties. As for other Riemannian computations on SPD manifolds, please refer to \cite{pennect2006aid,sra2011sdiv} for more studies on their properties.

\subsection{Other Layers}

After applying the LogEig layer, the vector forms of the outputs can be fed into classical Euclidean network layers. For example, the Euclidean fully connected (FC) layer could be inserted after the LogEig layer. The dimensionality of the filters in the FC layer is set to ${d_{k} \times d_{k-1}}$, where $d_{k}$ and $d_{k-1}$ are the class number and the dimensionality of the vector forms of the input matrices respectively. The final output layer for visual recognition tasks could be a softmax layer used in the context of Euclidean networks.

In addition, the pooling layers and the normalization layers are also important to improve regular Euclidean ConvNets. For the SPDNet, the pooling on SPD matrices can be first carried out on their matrix logarithms, and then transform them back to SPD matrices by employing the matrix exponential map $\exp(\cdot)$ in the Riemannian framework \cite{pennect2006aid,arsigny2007led}. Similarly, the normalization procedure on SPD matrices could be first to calculate the mean and variance of their matrix logarithms, and then normalize them with their mean and variance as done in \cite{ioffe2015batch}.

\section{Riemannian Matrix Backpropagation}
\label{backprop}

The model of the proposed SPDNet can be written as a series of successive function compositions $f= f^{(l)} \circ f^{(l-1)} \ldots \circ f^{(1)}$ with a parameter tuple $\bm{W} = (\bm{W}_l, \bm{W}_{l-1}, \ldots, \bm{W}_1)$, where $f^{(k)}$ is the function for the $k$-th layer, $\bm{W}_k$ is the weight parameter of the $k$-th layer and $l$ is the number of layers. The loss of the $k$-th layer could be denoted by a function as $L^{(k)}=\ell \circ f^{(l)} \circ \ldots f^{(k)}$, where $\ell$ is the loss function for the final output layer.

Training deep networks often uses stochastic gradient descent (SGD) algorithms. The key operation of one classical SGD algorithm is to compute the gradient of the objective function, which is obtained by an application of the chain rule known as backpropagation (backprop). For the $k$-th layer, the gradients of the weight $\bm{W}_k$ and the data $\bm{X}_{k-1}$  can be respectively computed by backprop as
\begin{alignat}{2}
	\frac{\partial L^{(k)}(\bm{X}_{k-1}, y)}{\partial \bm{W}_k} &  = \frac{\partial L^{(k+1)}(\bm{X}_k, y)}{\partial \bm{X}_k}\frac{ \partial f^{(k)}(\bm{X}_{k-1})}{\partial \bm{W}_k}, \label{Eq5} \\
	\frac{\partial L^{(k)}(\bm{X}_{k-1}, y)}{\partial \bm{X}_{k-1}}  &= \frac{\partial L^{(k+1)}(\bm{X}_k, y)}{\partial \bm{X}_k}\frac{ \partial f^{(k)}(\bm{X}_{k-1})}{\partial \bm{X}_{k-1}}, \label{Eq6}
\end{alignat}
where $y$ is the output, $\bm{X}_k=f^{(k)}(\bm{X}_{k-1})$. Eqn.\ref{Eq5} is the gradient for updating $\bm{W}_k$, while Eqn.\ref{Eq6} is to compute the gradients of the involved data in the layers below. For simplicity, we often replace $\partial L^{(k)}(\bm{X}_{k-1}, y)$ with $\partial L^{(k)}$ in the sequel. 

There exist two key issues for generalizing backprop to the context of the proposed Riemannian network for SPD matrices. The first one is updating the weights in the BiMap layers. As we force the weights to be on Stiefel manifolds, merely using Eqn.\ref{Eq5} to  compute their Euclidean gradients rather than Riemannian gradients in the procedure of backprop cannot yield valid orthogonal weights. While the gradients of the SPD matrices in the BiMap layers can be calculated by Eqn.\ref{Eq6} as usual, computing those with EIG decomposition in the layers of ReEig and LogEig has not been well-solved by the traditional backprop. Thus, it is the second key issue for training the proposed network.

To tackle the first issue, we propose a new way of updating the weights defined in Eqn.\ref{Eq1} for the BiMap layers by exploiting an SGD setting on Stiefel manifolds. The steepest descent direction for the corresponding loss function $L^{(k)}(\bm{X}_{k-1}, y)$ with respect to $\bm{W}_k$ on the Stiefel manifold is the Riemannian gradient $\tilde{\nabla} L^{(k)}_{\bm{W}_k}$. To obtain it, the normal component of the Euclidean gradient $\nabla L^{(k)}_{\bm{W}_k}$ is subtracted to generate the tangential component to the Stiefel manifold. Searching along the tangential direction takes the update in the tangent space of the Stiefel manifold. Then, such the update is mapped back to the Stiefel manifold with a retraction operation. For more details about the Stiefel geometry and retraction, readers are referred to \cite{edelman1998geometry} and \cite{absil2008optimization} (Page 45-48, 59). Formally, an update of the current weight $\bm{W}_k^{t}$ on the Stiefel manifold respects the form
\begin{alignat}{2}
	\tilde{\nabla} L^{(k)}_{\bm{W}_k^t}&=\nabla L^{(k)}_{\bm{W}_k^t}-\nabla L^{(k)}_{\bm{W}_k^t}(\bm{W}_k^t)^T\bm{W}_k^t, \label{Eq7} \\
	\bm{W}_k^{t+1} &= \Gamma(\bm{W}_k^{t}-\lambda\tilde{\nabla} L^{(k)}_{\bm{W}_k^t}), \label{Eq8}
\end{alignat}
where $\Gamma$ is the retraction operation, $\lambda$ is the learning rate, $\nabla L^{(k)}_{\bm{W}_k}(\bm{W}_k^t)^T\bm{W}_k^t$ is the normal component of the Euclidean gradient that can be computed by using Eqn.\ref{Eq5} as
\begin{equation}
\nabla L^{(k)}_{\bm{W}_k^t}=2\frac{\partial L^{(k+1)}}{\partial \bm{X}_k}\bm{W}_k^t\bm{X}_{k-1}.
\label{Eq9}
\end{equation}

As for the second issue, we exploit the matrix generalization of backprop studied in \cite{ionescu2015matrix} to compute the gradients of the involved SPD matrices in the ReEig and LogEig layers. In particular, let $\mathcal{F}$ be a function describing the variations of the upper layer variables with respect to the lower layer variables, i.e., $d\bm{X}_{k}=\mathcal{F}(d\bm{X}_{k-1})$. With the function $\mathcal{F}$, a new version of the chain rule Eqn.\ref{Eq6} for the matrix backprop is defined as
\begin{equation}
\frac{\partial L^{(k)}(\bm{X}_{k-1}, y)}{\partial \bm{X}_{k-1}}  = \mathcal{F}^{*} \left(\frac{\partial L^{(k+1)}(\bm{X}_k, y)}{\partial \bm{X}_k}\right),
\label{Eq10}
\end{equation}
where $\mathcal{F}^{*}$ is a non-linear adjoint operator of $\mathcal{F}$, i.e., $\bm{B}: \mathcal{F}(\bm{C})=\mathcal{F}^{*}(\bm{B}) : \bm{C}$, the operator $:$ is the matrix inner product with the property $\bm{B}:\bm{C} = Tr(\bm{B}^T\bm{C})$.

Actually, both of the two functions Eqn.\ref{Eq2} and Eqn.\ref{Eq4} for the ReEig and LogEig layers involve the EIG operation $\bm{X}_{k-1}=\bm{U}_{k-1}\bm{\Sigma}_{k-1}\bm{U}_{k-1}^T$ (note that, to increase the readability, we drop the layer indexes for  $\bm{U}_{k-1}$ and $\bm{\Sigma}_{k-1}$ in the sequel). Hence, we introduce a virtual layer ($k^{'}$ layer) for the EIG operation. Applying the new chain rule Eqn.\ref{Eq10} and its properties, the update rule for the data $\bm{X}_{k-1}$ is derived as 
\begin{equation}
\begin{aligned}
& \frac{\partial L^{(k)}}{\partial \bm{X}_{k-1}}: d\bm{X}_{k-1} \\
& =  \mathcal{F}^{*} \left(\frac{\partial L^{(k^{'})}}{\partial \bm{U}}\right): d\bm{X}_{k-1} + \mathcal{F}^{*} \left(\frac{\partial L^{(k^{'})}}{\partial \bm{\Sigma}}\right): d\bm{X}_{k-1} \\
&= \frac{\partial L^{(k^{'})}}{\partial \bm{U}} : \mathcal{F} \left(d\bm{X}_{k-1}\right) + \frac{\partial L^{(k^{'})}}{\partial \bm{\Sigma}} : \mathcal{F} \left(d\bm{X}_{k-1}\right) \\
&= \frac{\partial L^{(k^{'})}}{\partial \bm{U}} : d\bm{U}+ \frac{\partial L^{(k^{'})}}{\partial \bm{\Sigma}} : d\bm{\Sigma},
\end{aligned}
\label{Eq11}
\end{equation}
where the two variations $d\bm{U}$ and $d\bm{\Sigma}$ are derived by the variation of the EIG operation $d\bm{X}_{k-1}=d\bm{U}\bm{\Sigma}\bm{U}^T+ \bm{U}d\bm{\Sigma}\bm{U}^T+\bm{U}\bm{\Sigma}d\bm{U}^T$ as:
\begin{alignat}{2}
	d\bm{U} &=2\bm{U}(\bm{P}^T \circ (\bm{U}^T d\bm{X}_{k-1}\bm{U})_{sym}), \label{Eq12}\\
	d\bm{\Sigma} &=(\bm{U}^T d\bm{X}_{k-1}\bm{U})_{diag}, \label{Eq13}
\end{alignat}
where $\circ$ is the Hadamard product, $\bm{D}_{sym}=\frac{1}{2}(\bm{D}+\bm{D}^T)$, $\bm{D}_{diag}$ is $\bm{D}$ with all off-diagonal elements being 0 (note that we also use these two denotations in the following), $\bm{P}$ is calculated by operating on the eigenvalues $\sigma$ in $\bm{\Sigma}$:
\begin{align}
	\bm{P}(i,j) =
	\begin{cases}
		\frac{1}{\sigma_i-\sigma_j}, & i \neq j,\\
		0, & i = j.
	\end{cases}
	\label{Eq14}
\end{align}

For more details to derive Eqn.\ref{Eq12} and Eqn.\ref{Eq13}, please refer to \cite{ionescu2015matrix}.
Plugging Eqn.\ref{Eq12} and Eqn.\ref{Eq13} into Eqn.\ref{Eq11} and using the properties of the matrix inner product $:$ can derive the partial derivatives of the loss functions for the ReEig and LogEig layers:
\begin{equation}
\begin{aligned}
\frac{\partial L^{(k)}}{\partial \bm{X}_{k-1}} & = 2 \bm{U} \left(\bm{P}^T \circ \left( \bm{U}^T \frac{\partial L^{(k^{'})}}{\partial \bm{U}}\right)_{sym}  \right)\bm{U}^T \\
& +\bm{U} \left(\frac{\partial L^{(k^{'})}}{\partial \bm{\Sigma}}\right)_{diag}\bm{U}^T,
\label{Eq15}
\end{aligned}
\end{equation}
where $\frac{\partial L^{(k^{'})}}{\partial \bm{U}}$ and $\frac{\partial L^{(k^{'})}}{\partial \bm{\Sigma}}$ can be obtained with the same derivation strategy used in Eqn.\ref{Eq11}. For the function Eqn.\ref{Eq2} employed in the ReEig layers, its variation becomes $d\bm{X}_{k}=2(d\bm{U}\max(\epsilon\bm{I}, \bm{\Sigma})\bm{U}^T)_{sym}+ (\bm{U}\bm{Q}d\bm{\Sigma}\bm{U}^T)_{sym}$, and these two partial derivatives can be computed by
\begin{alignat}{2}
	\frac{\partial L^{(k^{'})}}{\partial \bm{U}} &= 2\left( \frac{\partial L^{(k+1)}}{\partial \bm{X}_k} \right)_{sym} \bm{U} \max(\epsilon\bm{I}, \bm{\Sigma}), \label{Eq16}\\
	\frac{\partial L^{(k^{'})}}{\partial \bm{\Sigma}} &= \bm{Q} \bm{U}^T \left(\frac{\partial L^{(k+1)}}{\partial \bm{X}_k} \right)_{sym} \bm{U}, \label{Eq17}
\end{alignat}
where $\max(\epsilon\bm{I}, \bm{\Sigma})$ is defined in Eqn.\ref{Eq3}, and $\bm{Q}$ is the gradient of $\max(\epsilon\bm{I}, \bm{\Sigma})$ with diagonal elements being defined as
\begin{align}
	\bm{Q}(i,i) =
	\begin{cases}
		1, & \bm{\Sigma}(i,i) > \epsilon,\\
		0, & \bm{\Sigma}(i,i) \leq \epsilon.
	\end{cases}\label{Eq18}
\end{align}

For the function Eqn.\ref{Eq4} used in the LogEig layers, its variation is $d\bm{X}_{k}=2(d\bm{U}\log(\bm{\Sigma})\bm{U}^T)_{sym}+ (\bm{U}\bm{\Sigma}^{-1}d\bm{\Sigma}\bm{U}^T)_{sym}$. Then we calculate the following two partial derivatives:
\begin{alignat}{2}
	\frac{\partial L^{(k^{'})}}{\partial \bm{U}} &= 2\left( \frac{\partial L^{(k+1)}}{\partial \bm{X}_k} \right)_{sym} \bm{U} \log(\bm{\Sigma}). \label{Eq19}\\
	\frac{\partial L^{(k^{'})}}{\partial \bm{\Sigma}} &= \bm{\Sigma}^{-1} \bm{U}^T \left( \frac{\partial L^{(k+1)}}{\partial \bm{X}_k} \right)_{sym} \bm{U}, \label{Eq20}
\end{alignat}

By mainly employing Eqn.\ref{Eq7}--Eqn.\ref{Eq9} and Eqn.\ref{Eq15}--Eqn.\ref{Eq20}, the Riemannian matrix backprop for training the SPDNet can be realized. The convergence analysis of the used SGD algorithm on Riemannian manifolds follows the developments in \cite{bottou2010large,bonnabel2013stochastic}.

\section{Discussion}

Although the two works \cite{harandi2014manifold,huang2015leml} have studied the geometry-aware map to preserve SPD structure, our SPDNet exploits more general (i.e., Stiefel manifold) setting for the map, and sets it up in the context of deep learning. Furthermore, we also introduce a non-linearity during learning SPD matrices in the network. Thus, the main theoretical advantage of the proposed SPDNet over the two works is its ability to perform deep learning and non-linear learning mechanisms.

While \cite{ionescu2015matrix} introduced a covariance pooling layer into ConvNets starting from images, our SPDNet works on SPD matrices directly and exploits multiple layers tailored for SPD matrix deep learning. From another point of the view, the proposed SPDNet can be built on the top of the network \cite{ionescu2015matrix} for deeper SPD matrix learning that starts from images. Moreover, we must claim that our SPDNet is still useful while \cite{ionescu2015matrix} will totally break down when the processed data are not covariance matrices for images.

\section{Experiments}
\label{others}

We evaluate the proposed SPDNet for three popular visual classification tasks including emotion recognition, action recognition and face verification, where SPD matrix representations have achieved great successes. The comparing state-of-the-art SPD matrix learning methods are Covariance Discriminative Learning (CDL) \cite{wang2012covariance}, Log-Euclidean Metric Learning (LEML) \cite{huang2015leml} and SPD Manifold Learning (SPDML) \cite{harandi2014manifold} that uses affine-invariant metric (AIM) \cite{pennect2006aid} and stein divergence \cite{sra2011sdiv}. The Riemannian Sparse Representation (RSR) \cite{harandi2012sparse} for SPD matrices is also evaluated. Besides, we measure the deep second-order pooling (DeepO2P) network \cite{ionescu2015matrix} which introduces a covariance pooling layer into typical ConvNets. For all of them, we use their source codes from authors with tuning their parameters according to the original works. For our SPDNet, we study 4 configurations, i.e., SPDNet-0BiRe/1BiRe/2BiRe/3BiRe, where $i$BiRe means using $i$ blocks of BiMap/ReEig. For example, the structure of SPDNet-3BiRe is $\bm{X}_0 \rightarrow f_b^{(1)} \rightarrow f_r^{(2)} \rightarrow f_b^{(3)} \rightarrow f_r^{(4)} \rightarrow f_b^{(5)} \rightarrow f_l^{(6)} \rightarrow f_f^{(7)} \rightarrow f_s^{(8)}$, where $f_b, f_r, f_l, f_f, f_s$ indicate the BiMap, ReEig, LogEig, FC and softmax log-loss layers respectively. The learning rate $\lambda$ is fixed as $10^{-2}$, the batch size is set to 30, the weights are initialized as random semi-orthogonal matrices, and the rectification threshold $\epsilon$ is set to $10^{-4}$. For training the SPDNet, we just use an i7-2600K (3.40GHz) PC without any GPUs.

\subsection{Emotion Recognition}

We use the popular Acted Facial Expression in Wild (AFEW) \cite{dhall2014emotion} dataset for emotion recognition. The AFEW database collects 1,345 videos of facial expressions of actors in movies with close-to-real-world scenarios.

The database is divided into training, validation and test data sets where each video is classified into one of seven expressions. Since the ground truth of the test set has not been released, we follow \cite{liu2014learning} to report the results on the validation set. To augment the training data, we segment the training videos into 1,747 small clips.
For the evaluation, each facial frame is normalized to an image of size $20 \times 20$. Then, following \cite{wang2012covariance}, we compute the covariance matrix of size $400 \times 400$ to represent each video.

On the AFEW database, the dimensionalities of the SPDNet-3BiRe transformation matrices are set to ${400 \times 200}$, ${200 \times 100}, {100 \times 50}$ respectively. Training the SPDNet-3BiRe per epoch (500 epoches in total) takes around 2 minutes(m) on this dataset.
As show in Table.\ref{tab1}, we report the performances of the competing methods including the state-of-the-art method (STM-ExpLet \cite{liu2014learning}) on this database. It shows our proposed SPDNet-3BiRe achieves several improvements over the state-of-the-art methods although the training data is small.

\begin{figure}
	\begin{center}
		\includegraphics[width=0.77\linewidth]{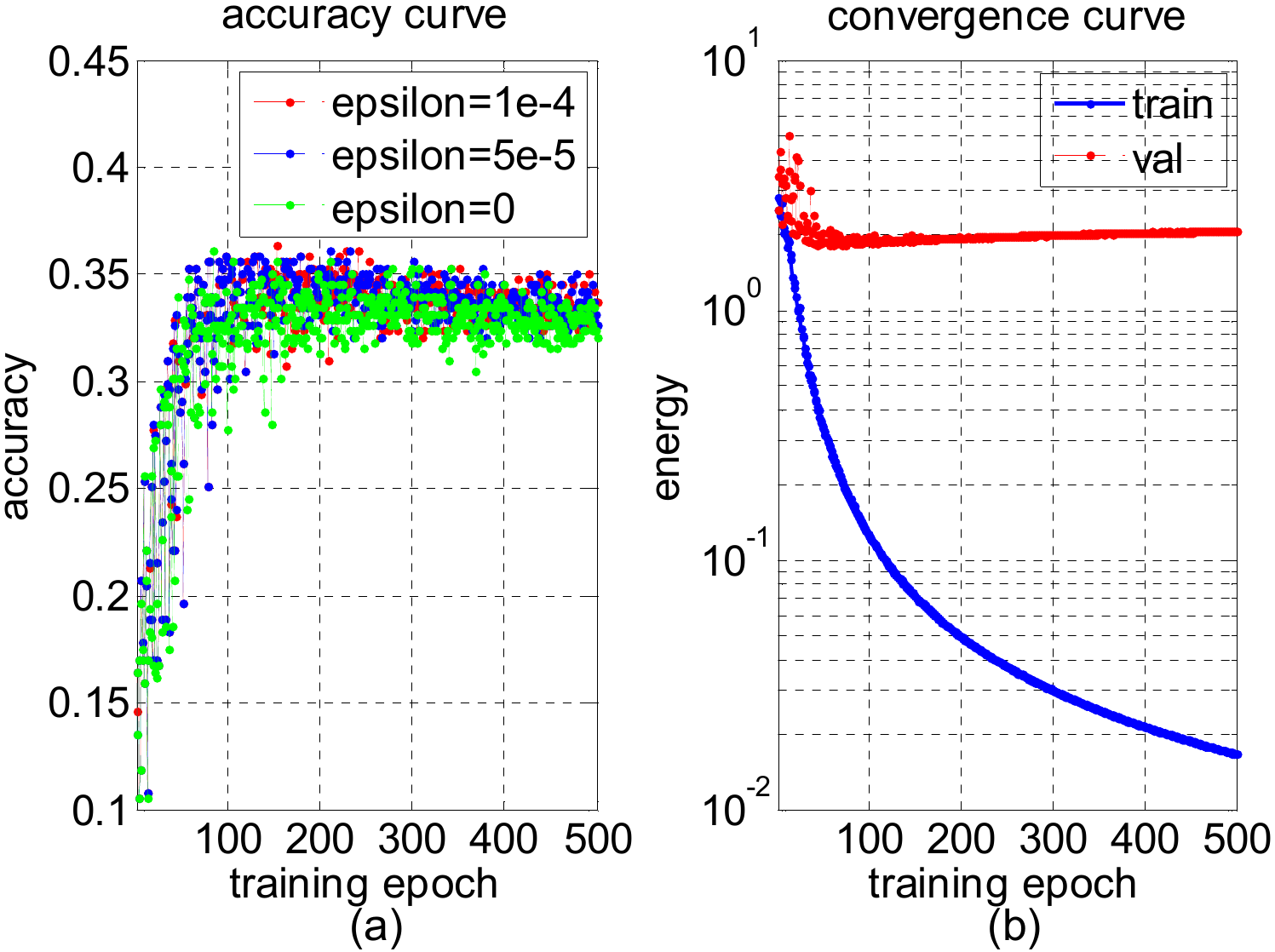}
		\caption{(a) Accuracy curve of the proposed SPDNet-3BiRe at different rectification threshold $\epsilon$ values, and (b) its convergence curve at $\epsilon=10^{-4}$ for the AFEW dataset.}
		\label{Fig2}
	\end{center}
\end{figure}

In addition, we study the behaviors of the proposed SPDNet with different settings. First, we evaluate our SPDNet without using the LogEig layer. Its extremely low accuracy 21.49\% shows the layer for Riemannian computing is necessary. Second, we study the case of learning directly on Log-Euclidean forms of original SPD matrices, i.e., SPDNet-0BiRe. The performance of SPDNet-0BiRe is 26.32\%, which is clearly outperformed by the deeper SPDNets (e.g., SPDNet-3BiRe). This justifies the importance of using the SPD layers. Besides, SPDNets also clearly outperform DeepO2P that inserts one LogEig-like layer into a standard ConvNet architecture. This somehow validates the improvements are from the contribution of the BiMap and ReEig layers rather than deeper architectures. Third, to study the gains of using multiple BiRe blocks, we compare SPDNet-1BiRe and SPDNet-2BiRe that feed the LogEig layer with SPD matrices of the same size as set in SPDNet-3BiRe. As reported in Table.\ref{tab1}, the performance of our SPDNet is improved when stacking more BiRes. Fourth, we also study the power of the designed ReEig layers in Fig.\ref{Fig2} (a). The accuracies of the three different rectification threshold settings $\epsilon=10^{-4}, 5 \times 10^{-5},0$ are $34.23\%, 33.15\%, 32.35\%$ respectively, that verifies the success of introducing the non-linearity. Lastly, we also show the convergence behavior of our SPDNet in Fig.\ref{Fig2} (b), which suggests it can converge well after hundreds of epoches.

\subsection{Action Recognition}

We handle the problem of skeleton-based human action recognition using the HDM05 database \cite{muller2007hdm}, which is one of the largest-scale datasets for the problem.

On the HDM05 dataset, we conduct 10 random evaluations, in each of which half of sequences are randomly selected for training, and the rest are used for testing. Note that the work \cite{harandi2014manifold} only recognizes 14 motion classes while the protocol designed by us is to identify 130 action classes and thus be more challenging. For data augmentation, the training sequences are divided into around 18,000 small sequences in each random evaluation.
As done in \cite{harandi2014manifold}, we represent each sequence by a joint covariance descriptor of size $93 \times 93$, which is computed by the second order statistics of the 3D coordinates of the 31 joints in each frame.

For our SPDNet-3BiRe, the sizes of the transformation matrices are set to ${93 \times 70}, {70 \times 50}, {50 \times 30}$ respectively, and its training time at each of 500 epoches is about 4m on average. Table.\ref{tab1} summarizes the results of the comparative algorithms and of the state-of-the-art method (RSR-SPDML) \cite{harandi2014manifold} on this dataset. As DeepO2P \cite{ionescu2015matrix} is merely for image based visual classification tasks, we do not evaluate it in the 3D skeleton based action recognition task. We find that our SPDNet-3BiRe outperforms the state-of-the-art shallow SPD matrix learning methods by a large margin (more than 13\%). This shows that the proposed non-linear deep learning scheme on SPD matrices leads to great improvements when the training data is large enough. 

The studies on without using LogEig layers and different configurations of BiRe blocks are executed as the way of the last evaluation. The performance of the case of without using LogEig layers is 4.89\%, again validating the importance of the Riemannian computing layers. Besides, as seen from Table.\ref{tab1}, the same conclusions as before for different settings of BiRe blocks can be drew on this database.

\begin{table}
	\begin{center}
		\footnotesize
		\begin{tabular}{|l|m{0.85cm}<{\centering}|m{1.65cm}<{\centering}|m{0.85cm}<{\centering}m{0.85cm}<{\centering}|}
			\hline
			Method & AFEW & HDM05 &  PaSC1 & PaSC2\\
			\hline\hline
			STM-ExpLet  & 31.73\% & -- & -- & -- \\
			RSR-SPDML  & 30.12\% & 48.01\%$\pm$3.38 & -- & --\\
			DeepO2P & 28.54\% & -- & 68.76\% & 60.14\%\\
			HERML-DeLF & -- & -- & 58.0\% & 59.0\%\\
			VGGDeepFace & -- & -- & 78.82\% & 68.24\%\\
			\hline
			CDL  & 31.81\% & 41.74\%$\pm$1.92 & 78.29\% & 70.41\% \\
			LEML  & 25.13\%  & 46.87\%$\pm$2.19 & 66.53\% & 58.34\%  \\
			SPDML-AIM  & 26.72\% & 47.25\%$\pm$2.78 & 65.47\% & 59.03\%   \\
			SPDML-Stein  & 24.55\% & 46.21\%$\pm$2.65 & 61.63\% & 56.67\%  \\
			RSR  & 27.49\% & 41.12\%$\pm$2.53 & -- & --\\
			\hline
			SPDNet-0BiRe & 26.32\% & 48.12\%$\pm$3.15 & 68.52\% & 63.92\% \\
			SPDNet-1BiRe & 29.12\% &  55.26\%$\pm$2.37 &  71.75\% & 65.81\% \\
			SPDNet-2BiRe & 31.54\% &  59.13\%$\pm$1.78 & 76.23\% & 69.64\% \\
			SPDNet-3BiRe & \textbf{34.23\%} & \textbf{61.45\%$\pm$1.12} & \textbf{80.12\%} & \textbf{72.83\%} \\
			\hline
		\end{tabular}
		\caption{The results for the AFEW, HDM05 and PaSC datasets. PaSC1/PaSC2 are the control/handheld testings.}
		\label{tab1}
	\end{center}
\end{table}

\subsection{Face Verification}

For face verification, we employ the Point-and-Shoot Challenge (PaSC) database \cite{beveridge2013challenge}, which is very challenge and widely-used for verifying faces in videos. It includes 1,401 videos taken by control cameras and 1,401 videos captured by handheld cameras for 265 people. In addition, it also contains 280 videos for training.

On the PaSC database, there are control and handheld face verification tasks, both of which are to verify a claimed \mbox{identity} in the query video by comparing with the associated target video. As done in \cite{beveridge2015report}, we also use the training data (COX) \cite{huang2015benchmark} with 900 videos. Similar to the last two experiments, the whole training data are also augmented to 12,529 small video clips. For evaluation, we use the approach of \cite{parkhi2015deep} to extract state-of-the-art deep face features on the normalized face images of size $224 \times 224$. To speed up the training, we employ PCA to finally get 400-dimensional features. As done in \cite{huang2015leml}, we compute an SPD matrix of size $401 \times 401$ for each video to fuse its data covariance matrix and mean.

For the evaluation, we configure the sizes of the SPDNet-3BiRe weights to ${401 \times 200}, {200 \times 100}, {100 \times 50}$ respectively. The time for training the SPDNet-3BiRe at each of 100 epoches is around 15m. Table.\ref{tab1} compares the accuracies of the different methods including the state-of-the-art methods (HERML-DeLF \cite{beveridge2015report} and VGGDeepFace \cite{parkhi2015deep}) on the PaSC database. Since the RSR method is designed for recognition tasks rather than verification tasks, we do not report its results. Although the used softmax output layer in our SPDNet is not favorable for the verification tasks, we find that it still achieves the highest performances.

Finally, we can also obtain the same conclusions as before for the studies on different configurations of the proposed SPDNet as observed from the results on the PaSC dataset.

\section{Conclusion}

We proposed a novel deep Riemannian network architecture for opening up a possibility of SPD matrix non-linear learning. To train the SPD network, we exploited a new backpropagation with an SGD setting on Stiefel manifolds. The evaluations on three visual classification tasks studied the effectiveness of the proposed network for SPD matrix learning.
As future work, we plan to explore more layers, e.g., parallel BiMap layers, pooling layers and normalization layers, to improve the Riemannian network. For further deepening the architecture, we would build the SPD network on the top of existing convolutional networks such as \cite{ionescu2015matrix} that start from images. In addition, other interesting directions would be to extend this work for a general Riemannian manifold or to compress traditional network architectures into more compact ones with the proposed SPD or orthogonal constraints.

\section{ Acknowledgments}
The authors gratefully acknowledge support by EU Framework Seven project ReMeDi (grant 610902).

\end{document}